\documentclass{bmcart}

\usepackage{amsthm,amsmath}
\usepackage{booktabs}
\usepackage{multicol}
\usepackage{multirow}
\usepackage{systeme}
\usepackage[utf8]{inputenc} 
\usepackage{graphicx}
\usepackage{subfig}
\graphicspath{{./images/}}

\startlocaldefs
\endlocaldefs

\begin{document}

\begin{frontmatter}

\begin{fmbox}
\dochead{Research}


\title{Leveraging a Joint of Phenotypic and Genetic Features on Cancer Patient Subgrouping}


\author[
   addressref={aff1},
   email={oniani.david@mayo.edu}
]{\inits{DO}\fnm{David} \snm{Oniani}}
\author[
  addressref={aff2},
]{\inits{CW}\fnm{Chen} \snm{Wang}}
\author[
  addressref={aff4},
]{\inits{YZ}\fnm{Yiqing} \snm{Zhao}}
\author[
  addressref={aff4},
]{\inits{WA}\fnm{Andrew} \snm{Wen}}
\author[
  addressref={aff4},
]{\inits{HL}\fnm{Hongfang} \snm{Liu}}
\author[
   addressref={aff4},
   corref={aff4},
   email={shen.feichen@mayo.edu}
]{\inits{FS}\fnm{Feichen} \snm{Shen}}


\address[id=aff1]{%
  \orgname{Kern Center for the Science of Health Care Delivery, Mayo Clinic},
  \city{Rochester},
  \cny{US}
}

\address[id=aff2]{%
  \orgname{Division of Biomedical Statistics and Informatics, Mayo Clinic},
  \city{Rochester},
  \cny{US}
}

\address[id=aff4]{%
  \orgname{Division of Digital Health Sciences, Mayo Clinic},
  \city{Rochester}
  \cny{US}
}



\end{fmbox}


\begin{abstractbox}

\begin{abstract} 

\parttitle{Background}
Cancer is responsible for millions of deaths worldwide every year. Although significant progress has been
achieved in cancer medicine, many issues remain to be addressed for improving cancer therapy. Appropriate cancer patient stratification is the prerequisite for selecting appropriate treatment plan, as cancer patients are of known heterogeneous genetic make-ups and phenotypic differences. In this study, built upon deep phenotypic characterizations extractable from Mayo Clinic electronic health records (EHRs) and genetic test reports for a collection of cancer patients, we developed a system leveraging a joint of phenotypic and genetic features for cancer patient subgrouping.

\parttitle{Methods}
The workflow is roughly divided into three parts: feature preprocessing, cancer patient classification, and cancer patient clustering based. In feature preprocessing step, we performed filtering, retaining the most relevant features. In cancer patient classification, we utilized joint categorical features to build a patient-feature matrix and applied nine different machine learning models, Random Forests (RF), Decision Tree (DT), Support Vector Machine (SVM), Naive Bayes (NB), Logistic Regression (LR), Multilayer Perceptron (MLP), Gradient Boosting (GB), Convolutional Neural Network (CNN), and Feedforward Neural Network (FNN), for classification purposes. Finally, in the cancer patient clustering step, we leveraged joint embeddings features and patient-feature associations to build an undirected feature graph and then trained the cancer feature node embeddings.

\parttitle{Results}
In cancer patient classification, the optimal performance of F1 0.9 was achieved by utilizing joint features. For cancer feature graph embeddings, average operation for edge embeddings yielded the highest ROC score of 0.71, followed by Hadamard. Regarding cancer patient clustering, CNN in combination with fine-tuned cancer feature graph embeddings (CFEmb+) showed the best purity score of 0.48, followed by UMLS/Mut2vec embeddings (UMEmb). In the qualitative evaluation, we observed some generated clusters had known relationships between diseases, phenotypes and genetic mutations.

\parttitle{Conclusions}
We utilized Mayo Clinic EHRs and genetic test reports in order to design a system for cancer patient subgrouping. We  discovered  the potential  usage of applying machine learning algorithms over joint phenotypic and genetic features for a better cancer type characterization.

\end{abstract}




\end{abstractbox}
%

\end{frontmatter}



\section{Introduction} \label{Introduction}
Cancer is a type of disease in which abnormal cells divide without control and can invade nearby tissues and other parts of the body via the blood and lymph systems \cite{yuan2016deepgene,feuerstein2007defining}. Nowadays, cancer is responsible for millions of deaths worldwide every year. Although significant progress has been achieved in cancer medicine, many issues remain to be addressed for improving cancer therapy. The goal of Precision Oncology is to enable oncologist practitioners to make better clinical decisions by incorporating individual cancer patients’ genetic information and clinical characteristics \cite{schwartzberg2017precision}. Precision Oncology is expected to improve selection of targeted therapies, tailored to individual patients and ultimately improve cancer patients’ outcomes. However, patient heterogeneity needs to be considered while providing cancer treatments. It is important to take patient subgroups into consideration in which the mechanism of diseases with the same group are more likely to be homogeneous \cite{on2014improving}. Therefore, appropriate cancer patient subgrouping is the prerequisite for selecting appropriate treatment plan, as cancer patients are of known heterogeneous genetic make-ups and phenotypic differences. 

Advances in cancer care have demonstrated the benefits of Individualized Medicine research, and the premise of Precision Oncology is to refer similar cancer outcomes to similar patients with similar features \cite{parimbelli2018patient}. Phenotypic characterizations are one of the key features to group patients by analyzing their clinical observations \cite{seligson2020recommendations}. Taking breast cancer as an example, specific sub-phenotypes have been leveraged to make decision making for therapies selection in practice \cite{lehmann2011identification}. Moreover, phenotypic heterogeneity amongs 7 subtypes of triple-negative breast cancer could be observed and analyzed \cite{masuda2013differential}. With the advance of machine learning techniques, computational deep phenotyping is widely used and described as the process to analyze phenotypic abnormalities precisely and comprehensively, in order to receive a better understanding of the natural history of a disease and its genotype-phenotype associations. For example, the DeepPhe information model is able to extract cancer-related phenotypes as categorical features by mining clinical narratives or structured datasets \cite{savova2017deepphe}. Beeghly-Fadiel et al. have extended DeepPhe to ovarian cancer \cite{beeghly2019deep}. Besides categorical phenotypic features, embeddings representations of phenotypes are also significant to deep phenotyping. HPO2Vec+ provides a way to learn node embeddings for the Human Phenotype Ontology (HPO) and be able to represent patients with vectorized representations for individual-level phenotypic similarity measurement \cite{shen2019hpo2vec+,shen2018constructing}. Maldonado et al. presented an approach that uses Generative Adversarial Networks (GANs) \cite{goodfellow2017generative} for learning UMLS embeddings and demonstrated the use case on clinical prediction tasks \cite{maldonado2019adversarial}.

In addition to phenotypic characterizations, features contained in genetic test reports provide crucial evidence for the decision making in practice. Genetic tests for targeted cancer therapy detect mutations in the DNA of cancer cells. Being aware of whether the cancer has a particular mutation can help guide the type of treatment. \cite{sawyers2004targeted}. Specifically, pathogenic variants demonstrate the genetic alteration that increases an individual's susceptibility or predisposition to a certain cancer type, and variant of uncertain significance (VUS) indicates a genetic variant that has been identified but without knowing too much details regarding significance to the function of an organism. There exist some studies that use machine learning to support variant interpretation in a clinical setting. For example, the LEAP model classifies variants based on features collected from functional predictions, splice predictions, clinical observation data and so on \cite{lai2020leap}. Some guidelines to define variant pathogenicity for somatic and VUS classification were established \cite{federici2020variants,richards2015standards,rehm2013acmg}. Mut2Vec provides a way to represent cancerous mutations with distributed embeddings, in order to support cancer classification and drug sensitivity prediction \cite{kim2018mut2vec}.

Although phenotypic and genetic features have their own advantages in facilitating computational individualized medicine and precision oncology, incorporating longitudinal clinical phenotypes into mutation features for cancer patient stratification is largely unexplored. In this study, built upon a cancer patient cohort with both deep phenothypic characterizations extractable from Mayo Clinic electronic health records (EHRs) and well-documented genetic test reports, we designed an approach leveraging a joint of phenotypic and genetic features for cancer patient subgrouping.

In the following, we first introduce materials and methods used in this study. We then describe experimental strategies followed by Results. We conclude our study with discussion and future work.

\section{Materials} \label{Materials}
We collected a cancer patient cohort (n=794) with accessible Mayo Clinic EHRs and foundation genetic test reports. Seven types of cancer are included in this cohort: lung cancer (n=223), prostate cancer (n=66), breast cancer (n=53), ovarian cancer (n=91), pancreas cancer (n=104), colon/rectum cancer (n=140), and liver cancer (n=107).

\subsection{Extract phenotypic features from the Mayo Clinic EHRs}
Given patient IDs retrieved from the cohort, we first applied an existing Mayo Clinic NLP-as-a-service implementation \cite{wen2019desiderata} to extract all the diagnostic-related terms from clinical notes and radiology report, normalized by the Unified Medical Language System (UMLS) \cite{bodenreider2004unified}. We then used a previously developed Human Phenotype Ontology (HPO)-based annotation pipeline \cite{shen2017phenotypic} to further map the UMLS terms to more specific HPO normalized phenotypic terms \cite{robinson2008human}. Specifically, for each cancer patient, we collected their phentoypic information within the last twelve months of the first mention of cancer diagnoses.

\subsection{Collect genetic feature from the FoundationOne CDx genetic test reports}
FoundationOne CDx is a next-generation sequencing (NGS) based assay that identifies genomic findings within hundreds of cancer-related genes. The foundationOne genetic test report is designed to provide physicians with clinically actionable information. The report for each patient is represented in the PDF format. We first used a python library named pdfminer to parse all the PDF report into text format, for the convenience of data processing. We then parsed out genes alteration and VUS sections for each patient. In particular, we treated mutations mined from the genes alteration section as pathogenic mutation, and we named mutations mined from the VUS section as VUS mutation. We used multi-hot encoding to represent genetic features for each patient, where 1 indicates the existence of pathogenic mutation or VUS, and 0 otherwise. 

\section{Methods} \label{Methods}
The workflow of this study is roughly divided into three parts. As shown in Figure \ref{figure1}, first of all, we preprocessed the extracted phenotypic and genetic features to remove noises. In addition, we adopted supervised learning strategy over the patients for cancer patient classification. Furthermore, we constructed a cancer feature graph based on associations between patients and different features, and we trained node embeddings based on the feature graph to get the vectorized feature representation. We then applied unsupervised learning approaches over the patient-level embeddings for patient clustering.

\subsection{Feature preprocessing}
For phenotypes, we first removed some general phenotypic terms mined from EHRs, including cyst, pain, carcinoma, neoplasm, symptoms, disease, minor (disease), and sarcoma - cateogry (morphologic abnormality). We then filtered out phenotypic terms with low frequency across all the patients, in order to largely avoid noises. By heuristics, we set the frequency threshold for selecting phenotypes as 20. Similarly, for genetic features, we only kept the ones that appear more than 10 times across all the patients.
\subsection{Cancer patient classification based on joint categorical features}
We built a \(M*N\) patient-feature matrix with multi-hot encoding, where \(M\) indicates the number of patients and \(N\) denotes the total number of features, including 239 phenotypic features, 328 genetic features, and 1 demographic feature for gender. Given patients' cancer diagnosis as labels, we treated cancer patient classification as a multi-class classification task and applied different supervised learning classifiers over the patient-feature matrix to categorize patients based on the diagnosed cancer types. Here, we used Random Forests (RF), Decision Tree (DT), Support Vector Machine (SVM), Naive Bayes (NB), Logistic Regression (LR), Multilayer Perceptron (MLP), Gradient Boosting (GB), Convolutional Neural Network (CNN), and Feedforward Neural Network (FNN).

\subsection{Cancer patient clustering based on joint embeddings features}

\subsubsection{Construction of the cancer feature graph}

As shown in Figure \ref{figure2}, associations between the patients and the features were captured to build the feature graph. Specifically, we made a link between phenotypic feature and genetic feature if both of them are associated with the same patient. As a result, the feature graph can be defined as \(G=(N,L)\), where \(N\) represents the set of phenotypic/genetic nodes and \(L\)
represents the set of links/edges between these nodes. We viewed graph \(G\) as an undirected graph that bridges mutations and phenotypes based on information collected from heterogeneous cancer patients in clinical practice.

\subsubsection{Train cancer feature node embeddings}
We proceed by training node embeddings for feature node contained in the cancer feature graph  \(G\).

Due to its demonstrated high performance on link prediction, we applied
node2vec model \cite{grover2016node2vec}. It implements a second-order random walk over the topological
structure of the graph. In every walk, there are three types of nodes involved.
Specifically, source, intermediate, and target nodes which in these case all
are phenotypic/genetic nodes.

In graph \(G\), given any source entity node as \(E_s\), target entity node as \(E_t\), we denote the
intermediate entity that exists on the path between \(E_s\) and \(E_t\) as
\(E_{st}\). Then the distribution  of  entity \(E_t\) with a fixed length
of random walk can be represented as follows: 
\begin{equation}
    P(E_t \mid E_i) = \systeme*{\dfrac{\pi(E_i, E_t)}{Z}\ if\ (E_i\, E_t)\ is\ a\ link, 0\ otherwise}
\end{equation}

where \(Z\) is the normalization constant and \(\pi(E_i, E_t)\) is the transition
probability between the phenotypic/genetic nodes \(E_i\) and \(E_t\).

We denote the weight over the link \((E_i, E_t)\) as \(w_(E_i, E_t) = 1\).
Then \(\pi(E_i, E_t)\) can be calculated as follows:

\begin{equation}
    \pi(E_i, E_t) = \alpha(E_s, E_t) \cdot w(E_i, E_t)
\end{equation}

where \(\alpha(E_s, E_t)\) is a search bias term and as defined in node2vec,
can be computed:

\begin{equation}
    \alpha(N_s, N_t) = \systeme*{\dfrac{1}{q}\ if\ D = 2, 1\ if\ D = 1, \dfrac{1}{p}\ if\ D = 0}
\end{equation}

where \(D\) is the shortest distance between the nodes \(E_s\) and \(E_t\)
and \(p\) and \(q\) are hyperparameters that balance between the breadth-first
search and depth-first search searching strategies for both local and global
optimization.

After learning the sampled network data using random walk, we then leveraged
the Skip-Gram model to train entity representations on the sampled data. For
each entity node\(E_s \in N\)and all its sampled  neighbors \(SN(E_s)\), the
loss function for entity representation learning could be described as shown in Equation (4). This loss function via Stochastic gradient descent. The prediction distribution was normalized using softmax function.

\begin{equation}
    \underset{f}{\max} = \sum_{E_s \in N} \log P(SN(E_s) \mid f(E_s))
\end{equation}

As a result, we have mapped the bipartite cancer feature graph from graph space to embeddings space. Each node is represented by a trained vector and strength of association for each pair of node is measurable. 

\subsubsection{Patient-level embeddings representation}
For any patient \(p\), the final patient level embeddings with respect to phenotype or genetic features will be represented by using average embeddings (AE): 
\begin{equation}
    AE(p) = \dfrac{\sum_{me_p \in pv_p}^{|pv_p|} f_c(me_p)}{|pv_p|}
\end{equation}

where \(f_c\) a feature matrix to map any phenotype or mutation \(me_p\) contained in patient vector \(pv_p\) to its corresponding embeddings, and \(c\) indicates which graph embeddings to be used (either our trained embeddings or existing ones). Given the initial patient embeddings with respect to phenotype as \(R_p\), patient embeddings with respect to genetic features as \(R_g\), and demographic multi-hot vector as \(R_d\), the final patient vector \(R_f\) will be represented by concatenating three vectors together as \([R_p, R_g, R_d]\)

\subsubsection{Patient clustering}

We further fine-tuned the initial patient-level embeddings using the cancer classification task via a categorical cross-entropy loss. We then adopted the t-distributed stochastic neighbor embedding (t-SNE) algorithm to render the refined patient embeddings for all patients into a 2D space \cite{maaten2008visualizing}. We first visualized patient subgrouping intuitively with t-SNE. As t-SNE does not perform clustering in and of itself, but instead renders each patient embedding into a \((x,y)\) coordinate. As such, additional clustering approach is required to re-categorize these points into discrete clusters. K-means clustering \cite{macqueen1967some} was therefore applied over patients to further partition different patient groups into distinct clusters.

\section{Evaluation} \label{Evaluation}

\subsection{Evaluate supervised learning performance}
We used 70\% data for training purpose and the rest for testing. We conducted a 10-fold cross-validation approach over the training data for choosing the optimal hyper-parameters for each classifier. To handle data imbalanced problem, we oversampled the training data. In order to demonstrate the advantage of combining both phenotypic and genetic features for cancer patient classification, we compared the cancer patient classification performance by using phenotypic feature only, genetic feature only, and the joint of two type of features. We used precision, recall, and F1-score to evaluate the overall classification performance and which for each specific cancer.

\subsection{Evaluate the construction of node embeddings}
We leveraged a link prediction task to train node embeddings for the cancer feature graph. We used 60\% of edges as training data, 10\% edges as validation set, and the rest is used for testing purpose. Similar to what reported in the node2vec study, we tested edge embeddings with four operations, including Hadamard, Average, L1, and L2 \cite{grover2016node2vec}. We combined RF with different edge embeddings to train the optimal node embedding through the link prediction task, and we reported ROC score for each experiment.

\subsection{Evaluate unsupervised learning performance}
We compared the trained cancer feature graph embeddings with other existing embeddings on cancer patient clustering. Specifically, UMLS embeddings \cite{maldonado2019adversarial} and Mut2vec \cite{kim2018mut2vec} are two pre-trained embeddings for representing phenotype and genetic mutation respectively. In this evaluation, according to Equation (5), we used features contained in UMLS embeddings and Mut2vec to represent patient-level embeddings, and then compared patient subgrouping performance generated by six experimental settings: original cancer feature graph embeddings (CFEmb), fine-tuned cancer feature graph embeddings with CNN (CFEmb+ with CNN), fine-tuned cancer feature graph embeddings with FNN (CFEmb+ with FNN), original UMLS/Mut2vec embeddings (UMEmb), fine-tuned UMLS/Mut2vec embeddings with CNN (UMEmb+ with CNN), and fine-tuned UMLS/Mut2vec embeddings with FNN (UMEmb+ with FNN). Size of dimension of the pre-trained UMLS embeddings and Mut2vec are 51 and 300 respectively. Since UMLS embeddings and Mut2vec don't encode gender as a feature, we used a 1-bit binary indicator to represent gender here. Therefore, the total dimension size for patient-level embeddings represented by UMLS embeddings and Mut2vec is 352. We visualized the t-SNE plots and selected the optimal number of clusters (ranging from 2 to 20) by measuring the purity scores for K-means clustering across all the experiments. We also conducted a qualitative analysis for clusters generated by the experiment with the optimal purity score. We picked the embeddings with the optimal purity score and analyzed the most frequent cancers, phenotypes, and genetics involved in some selected clusters.

\section{Results} \label{Results}
\subsection{Cancer patient classification}
As shown in Table \ref{table1}, for CNN, FNN, LR, MLP, RF, GB, and SVM, performance using joint features outpeformed the usage of phenotypic and genetic features alone respectively. We observed that phenotypic features collected from EHRs are able to better differentiate patients than genetic features collected from genetic test report in this study. In particular, for DT and NB, using phenotypic features alone outperformed using the combination of features. For joint features,  LR and MLP outperformed the rest of the models, with LR showing marginal F1-score advantage over MLP. FNN, RF, GB, and SVM also performed similarly with the performance slightly lower than both LR and MLP. NB classifier showed the worst performance with the precision, recall, and F1 values of 0.48, 0.40, and 0.40 respectively. CNN showed the second worst
performance with joint multi-hot encoding features. Its precision value was 0.77, recall was 0.73, and F1 score was 0.74.

\begin{table}[h]
    \begin{center}
        \caption{Comparison of overall cancer patient classification performance
                 between different features. The optimal precision, recall and F1 scores are in bold.}
        \label{table1}
        \begin{tabular}{cccccccccc}
            \toprule
            Model & \multicolumn{3}{c}{Joint features} & \multicolumn{3}{c}{Phenotypic features} & \multicolumn{3}{c}{Genetic features}\\
            \midrule
            & Precision & Recall & F1 Score & Precision & Recall & F1 Score & Precision & Recall & F1 Score\\
            \midrule
            CNN & \textbf{0.77} & \textbf{0.73} & \textbf{0.74} &          0.75  &          0.74  &         0.73        & 0.24      &          0.37     &         0.25\\
            \midrule
            FNN &        \textbf{0.88}  &         \textbf{0.86}  &         \textbf{0.87}  &          0.85  &          \textbf{0.86}  &         0.85& 0.15  &          0.19  &         0.17\\
            \midrule
            DT  &         0.76  &         \textbf{0.78}  &         0.76  &          \textbf{0.83}  &          0.77  &         \textbf{0.77} & 0.15  &          0.19  &         0.17\\
            \midrule
            LR  &         \textbf{0.90} &          \textbf{0.90}  &         \textbf{0.90}  &          0.84  &          0.84  &         0.83& 0.21  &          0.37  &         0.25\\
            \midrule
            MLP &         \textbf{0.89} &          \textbf{0.88}  &         \textbf{0.88}  &          0.83  &  0.82 & 0.82 & 0.19  &          0.36  &         0.23\\
            \midrule
            NB  &         0.48  &         0.40  &         0.40  &          \textbf{0.52}  &          \textbf{0.54}  &         \textbf{0.51}& 0.24  &          0.29  &         0.25\\
            \midrule
            RF  &         \textbf{0.87}  &         \textbf{0.87}  &         \textbf{0.86}  &  0.85 &  0.85 & 0.83 & 0.19  &          0.25  &         0.21\\
            \midrule
            GB  &         \textbf{0.89}  &         \textbf{0.85}  &         \textbf{0.86}  &  0.86 &  \textbf{0.85} & 0.85 & 0.19  &          0.24  &         0.21\\
            \midrule
            SVM &         \textbf{0.90}  &         \textbf{0.86}  &         \textbf{0.87}  &          0.87  &          \textbf{0.86}  &         0.85& 0.21  &          0.25  &         0.22\\
            \bottomrule
        \end{tabular}
    \end{center}
\end{table}

Table \ref{table2} is in accordance with the overall
results shown in Table \ref{table1}. The average F1 score for the
classification of ovarian cancer was the highest, equal to 0.85. Liver
cancer came next, with the average F1 score of 0.81. Average F1 score for the breast cancer classification was
0.78. Prostate cancer and pancreas cancer had the average F1 scores of 0.77 and 0.74
respectively. Lung cancer had the average F1 score of 0.73 and NB was not able to make any correct classification for lung cancer patients. Colon/rectum
cancer had the worst F1 score across all models used with joint
feature representations, with the score of 0.72.

\begin{table}[h]
    \begin{center}
        \caption{Classification F1 score on each cancer type using joint features. The optimal F1 scores are in bold.}
        \label{table2}
        \begin{tabular}{cccccccc}
            \toprule
            Model & Lung & Prostate & Breast & Ovarian & Pancreas & Col./Rec. & Liver\\
            \midrule
            CNN & 0.75 & 0.85 & 0.76 & 0.85 & 0.76 & 0.75 & 0.87\\
            \midrule
            DNN &         0.67  &         0.68  &          0.45 &          0.81 &         0.60  &         0.55  &         0.68\\
            \midrule
            DT  &         0.67  &         0.62  &          0.83 &          0.81 &         0.75  &         0.70  &         0.83\\
            \midrule
            LR  &         0.89  &         0.92  &          0.88 &          \textbf{0.93} &         \textbf{0.88}  &         \textbf{0.88}  &         0.91\\
            \midrule
            MLP &         0.89  &         \textbf{0.93}  &          0.82 &          \textbf{0.93} &         \textbf{0.88}  &         0.81  &         0.91\\
            \midrule
            NB  &         0.00  &         0.43  &          0.60 &          0.61 &         0.55  &         0.25  &         0.37\\
            \midrule
            RF  &         0.84  &         0.87  &          0.88 &          0.92 &         0.75  &         0.86  &         0.90\\
            \midrule
            GB  &         0.89  &         0.86  &          0.87 &          0.87 &         0.74  &         0.86  &         \textbf{0.93}\\
            \midrule
            SVM &         \textbf{0.94}  &         0.81  &          \textbf{0.89} &          \textbf{0.93} &         0.77  &         0.84  &         0.91\\
            \bottomrule
        \end{tabular}
    \end{center}
\end{table}

\subsection{Cancer feature graph embeddings}
As shown in Figure \ref{figure3}, link prediction performance made by RF with four edge embeddings operations were compared. We found that Average edge embeddings yielded the highest ROC score of 0.71, followed by Hadamard (area = 0.70). L1 and L2 had the same prediction performance of 0.63, respectively. Therefore, we used RF with Average edge embeddings to train the embeddings and the final hyper-parameters were selected as follows: return parameter \(p\) is 0.5, in-out parameter \(q\) is 0.5, window size is 10, number of walk is 5, walk length is 10, dimension = 100.

\subsection{Cancer patient clustering}
We first conducted a qualitative analysis of the patient-level clustering. Six t-SNE clustering visualizations were made as shown in Figure \ref{figure4}. Regarding the usage of our self-trained cancer feature graph embeddings, we observed that the original CFEmb was not able to subgroup cancer patients clearly based on the joint embeddings features. However, we found that fine-tuned CFEMb with both CNN and FNN were able to better organize patients with same cancer types. Particularly, there is no significant differences between CFEmb+ with CNN and CFEmb+ with FNN could be observed from visualizations. Intuitively, we found that refined features from both CFEmb+ can detect clear clusters for patients with prostate cancer (blue), breast cancer (green), ovarian cancer (black), and liver cancer (red). For the combination usage of UMLS embeddings and Mut2vec, we observed that there exists no significant differences on clustering cancer patients among the original UMEmb, UMEmb+ with CNN, and UMEmb+ with FNN. Compared to CFEmb+, UMEmb+ are less capable of making cohesion clusters to represent each type of cancer. Moreover, since CFEmb+ encoded gender as embeddings, we found that both CFEmb+ with CNN and FNN were able to make a clear separation for patients with different genders. For example, for CFEmb+ with CNN, all the male patients were grouped in the left part of the visualization (e.g., patients with prostate cancer (blue)) while all the female patients were grouped in the right part, including patients with breast cancer (green) and ovarian cancer (black). The same separation could be observed in CFEmb+ with FNN. While neither UMEmb+ with CNN nor UMEmb+ with FNN were capable of separating patients by gender.

In addition, we made a quantitative analysis for six embeddings using K-means. Table 3 depicts the number of clusters along with the associated best purity score selected for each embeddings. We observed that CFEmb+ with CNN conbritued to the optimal purity score of 0.48 while \(k=18\).

\begin{table}[h]
    \begin{center}
        \caption{Clustering performance over different patient-level embeddings. The optimal purity score is in bold.}
        \label{table3}
        \begin{tabular}{ccc}
        \toprule
            Methods & Number of clusters (k) & Purity score\\
            \midrule
            CFEmb & 12 & 0.30\\
             \midrule
            CFEmb+ with FNN & 19 & 0.39\\
             \midrule
            CFEmb+ with CNN & 18 & \textbf{0.48}\\
             \midrule
            UMEmb & 18 & 0.41\\
              \midrule
            UMEmb+ with FNN & 18 & 0.39\\
              \midrule
            UMEmb+ with CNN & 17 & 0.39\\
         \bottomrule
        \end{tabular}
    \end{center}
\end{table}

We picked CFEmb+ with CNN as the optimal embeddings and further analyzed some selected clusters. We observed that several generated clusters had known relationships between diseases, phenotypes and genetic mutations. For example, as shown in Table \ref{table4}, cluster-8 was mostly enriched with prostate cancer (n=39) and related disease phenotypes (prostatic disease, carcinoma in situ of prostate),  as well as TP53 pathogenic mutations known implicated in prostate cancer progression \cite{ecke2010tp53}; in addition, pathogenic alterations of androgen receptor (AR) is found associated with cluster-8, and has been clearly established as driver genetic event leading to prostate cancer and considered as target for various therapeutic options for treatment prostate cancer \cite{fujita2019role}. As two other examples, cluster-7 and cluster-3 were both enriched with colon/rectum cancers (n=24 and 21 respectively) and closely related phenotypes (e.g. colonic diseases); in addition, commonly mutated TP53 and KRAS genes were found in both clusters, with KRAS alternation as an established treatment-response marker and potential drug targets \cite{siddiqui2010kras, mccormick2015kras}. Interestingly, cluster-7 and cluster-3 lead to uniquely different VUS findings in BRCA2 (cluster-7) and TSC2 (cluster-3), raising the scientific possibilities that these colorectal disease were from different disease etiologies and mutational background. Although these observations require larger sample-size and additional cohort for independent validations, data-driven discovery procedure as such clearly pave a way for formulating novel hypothesis through joint mining of cancer, phenotypes and genetic changes.

\begin{table}[h]
\begin{center}
     \caption{Selected clusters with top most frequent cancer type, phenotypic features, and genetic features. Number in bracket indicates the frequency of entity. "-PATH" indicates a pathogenic mutation, while "-VUS" indicates a variant of uncertain significance.}
    \label{table4}
        \begin{tabular}{llll}
        \toprule
            Cluster ID & Cancer type & Phenotypic features & Genetic features\\
             \midrule
            \multirow{4}{*}{Cluster-8} 
                                 & \multicolumn{1}{l}{Prostate cancer (39)} & \multicolumn{1}{l}{Prostatic diseases (44)} & \multicolumn{1}{l}{TP53-PATH (18)}\\\cmidrule{2-4}
                          
                                    & \multicolumn{1}{l}{Colon/Rectum cancer (3)} & \multicolumn{1}{l}{Carcinoma in situ of prostate (44)} & \multicolumn{1}{l}{PTEN-PATH (17)}\\\cmidrule{2-4}
                                    & \multicolumn{1}{l}{Lung cancer (2)} & \multicolumn{1}{l}{Neoplasm of uncertain or unknown behavior of prostate (44)} & \multicolumn{1}{l}{TAF1-VUS (18)}\\\cmidrule{2-4}
                                        & \multicolumn{1}{l}{Pancreas cancer (2)} & \multicolumn{1}{l}{Benign neoplasm of prostate (44)} & \multicolumn{1}{l}{AR-PATH (18)}\\\cmidrule{1-4} 
                                        
            \multirow{4}{*}{Cluster-7} & \multicolumn{1}{l}{Colon/Rectum cancer (24)} & \multicolumn{1}{l}{Neoplasm of uncertain or unknown behavior of colon (35)} & \multicolumn{1}{l}{TP53-PATH (20)} \\\cmidrule{2-4}
                                 & \multicolumn{1}{l}{Lung cancer (7)} & \multicolumn{1}{l}{Colonic diseases (35)}& \multicolumn{1}{l}{KRAS-PATH (20)} \\\cmidrule{2-4}
                                    & \multicolumn{1}{l}{Pancreas cancer (3)} & \multicolumn{1}{l}{Stage 0 carcinoma of colon (35)}& \multicolumn{1}{l}{ASXL1-VUS (18)} \\\cmidrule{2-4}
                                 & \multicolumn{1}{l}{Liver cancer (1)} & \multicolumn{1}{l}{Benign neoplasm of liver (28)}& \multicolumn{1}{l}{BRCA2-VUS (7)} \\\cmidrule{1-4}
                          
            \multirow{4}{*}{Cluster-3} & \multicolumn{1}{l}{Colon/Rectum cancer (21)} & \multicolumn{1}{l}{Neoplasm of uncertain or unknown behavior of colon (34)} & \multicolumn{1}{l}{TP53-PATH (33)} \\\cmidrule{2-4}
                                 & \multicolumn{1}{l}{Lung cancer (6)} & \multicolumn{1}{l}{Colonic diseases (34)}& \multicolumn{1}{l}{ASXL1-VUS (18)} \\\cmidrule{2-4}
                                    & \multicolumn{1}{l}{Ovarian cancer (4)} & \multicolumn{1}{l}{Stage 0 carcinoma of colon (34)}& \multicolumn{1}{l}{KRAS-PATH (15)} \\\cmidrule{2-4}
                                 & \multicolumn{1}{l}{Pancreas cancer (4)} & \multicolumn{1}{l}{Urologic diseases (30)}& \multicolumn{1}{l}{TSC2-VUS (13)} \\
         \bottomrule
        \end{tabular}
       

    \label{tab:xxx}
         \end{center}
\end{table}

\section{Discussion and Future Work} \label{Results}

For cancer patient classification with joint categorical features, we found that CNN was not able to outperform FNN and other conventional machine learning algorithms. The reason behind this is that the size of cancer patient cohort is limited and the CNN models trained on the small dataset usually yield poor generalization and horrible performance \cite{zhang2015initialize}. In addition, we observed that NB with phenotypic features outperformed which with joint features, despite the low F1 score of 0.51. Specifically, we used the Gaussian-based NB in evaluation and we measured average feature importance score using permutation feature importance measurement. We found that NB achieved an average score of 0.0008 and 0.0023 by using joint features and phenotypic features respectivley, indicating that homogeneous features were more suitable for NB in this evaluation.

Regarding cancer feature graph generation, there is a debating between whether to include patient as a graph node or not. We decided to not include patients as nodes but only use them for bridging heterogeneous phenotypic and genetic features, and here are rationales to implement it this way: 1) our plan is to build a generic embeddings regarding important associations between phenotypes and mutations. The constructed embeddings incorporates rich semantic information contained in both Mayo Clinic EHRs and FoundationOne CDx genetic test reports, which could be used as a pre-trained cancer feature embeddings for othe datasets and studies independently; 2) Since we adopted the 2nd order random walk within node2vec to train embeddings, adding patients as nodes in between phenotypic and genetic features will reduce the chance of reachability between phenotype and mutation nodes, thus weaken the power of association prediction/detection amongs features.

For cancer patient clustering, in Table \ref{table4}, we found that some most frequent phenotypic features have coarse granularity. For example, in Cluster-8, all the top 4 phenotypes contain keyword prostate or prostatic, that explains why prostate cancer is the most frequent cancer type in this cluster. In the future, we will explore more over phenotypic features with finer granularity, in order to increase the model prediction power while dealing with phenotypic features without any keyword mentions. To achieve this, we will focus on the semantic relationship amongst synonym, hyponymy, and hypernymy within the HPO. Our prior designed HPO2Vec+ framework \cite{shen2019hpo2vec+} will also be leveraged to capture strong semantic relationships for this purpose.

\section{Conclusion} \label{Results}

We utilized Mayo Clinic electronic health records (EHRs) and genetic test reports in order to design a system for cancer patient subgrouping. For this purpose, we developed a three-step process comprised of feature preprocessing, cancer patient classification, and cancer patient clustering. We discovered that the usage of the joint phenotypic and genetic features showed a significant performance increase over the either set of features used alone. The usage of genetic features alone showed the worst performance. In this study, we discovered the potential usage of applying machine learning algorithms over joint phenotypic and genetic features for a better cancer type characterization.



\begin{backmatter}

\section*{Acknowledgements}
This work has been supported by the Gerstner Family Foundation, Center for Individualized Medicine (CIM) of Mayo Clinic, and Genentech Research Fund in Individualized Medicine.


\section*{Abbreviations}
EHRs: Electronic Health Records; RF: Random Forests; DT: Decision Tree; SVM: Support Vector Machine; NB: Naive Bayes; LR: Logistic Regression; MLP: Multilayer Perceptron; GB: Gradient Boosting; CNN: Convolutional Neural Network; FNN: FeedforwardNeural Network; CFEmb: Cancer Feature Graph Embeddings; CFEmb+: Fine-tuned Cancer Feature Graph Embeddings (CFEMb+); UMEmb: UMLS/Mut2vec embeddings; UMEmb+: Fine-tuned UMLS/Mut2vec embeddings.

\section*{Availability of data and materials}
The cancer network file is stored in .txt format and the cancer network embeddings file is saved in .emb format. Location of the data is at: https://github.com/shenfc/Joint-Learning. Data files could be downloaded freely from GitHub.

\section*{Ethics approval and consent to participate}
This study used existing data to conduct a retrospective study. The study
and a waiver of informed consent were approved by Mayo Clinic Institutional
Review Board in accordance with 45 CFR 46.116 (Approval \#17–003030, \#20-001474)All methods were carried out in accordance with relevant guidelines and regulations.

\section*{Competing interests}
The authors declare that they have no competing interests.

\section*{Consent for publication}
The author(s) declare(s) that the manuscript does not contain any individual
person’s data. So this paper requires no consent to publish.

\section*{Authors' contributions}
All co-authors are justifiably credited with authorship, according to the authorship criteria. Final approval is given by each co-author. In detail: DO contributed to the system implementation, data analysis, result interpretation, and manuscript preparation. CW contributed to system design, result interpretation, and manuscript preparation. YZ contributed to data extraction and data cleaning. AW is responsible for building data engineering pipeline and extracting phenotypic data from electronic health records. HL contributed to system design and evaluation. FS conceptualized and led the study, contributed to system design and implementation, data analysis, results interpretation, and manuscript preparation.



\bibliographystyle{bmc-mathphys} 
\bibliography{bmc_article}      




\section*{Figures}

  \includegraphics[width=15cm]{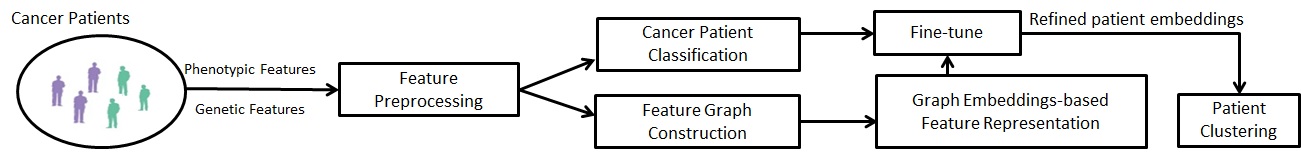}
  \centering Figure 1. The overall workflow
  caption{The overall workflow}
  \label{figure1}

\includegraphics[width=15cm]{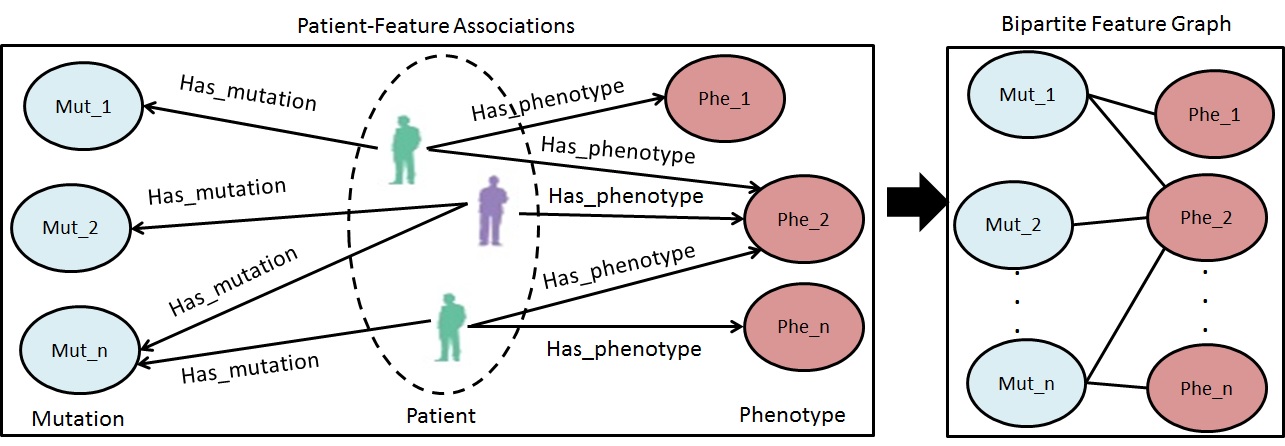}
\centering Figure 2. Building cancer feature graph
    caption{Building cancer feature graph}
    \label{figure2}

\includegraphics[width=15cm]{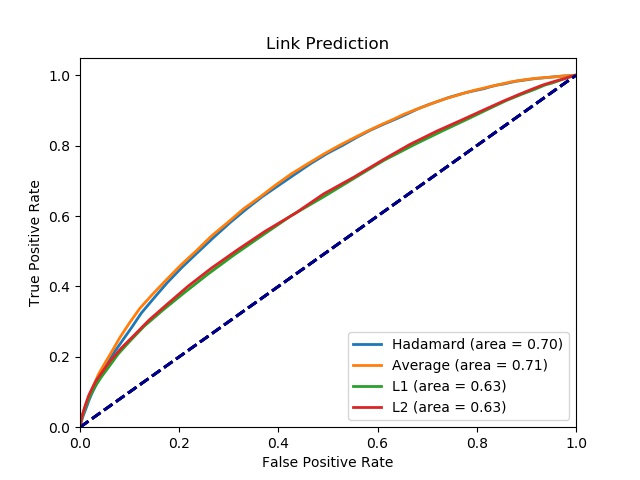}
\centering Figure 3. ROC for link prediction
    caption{ROC for link prediction}
    \label{figure3}
    
\includegraphics[width=15cm]{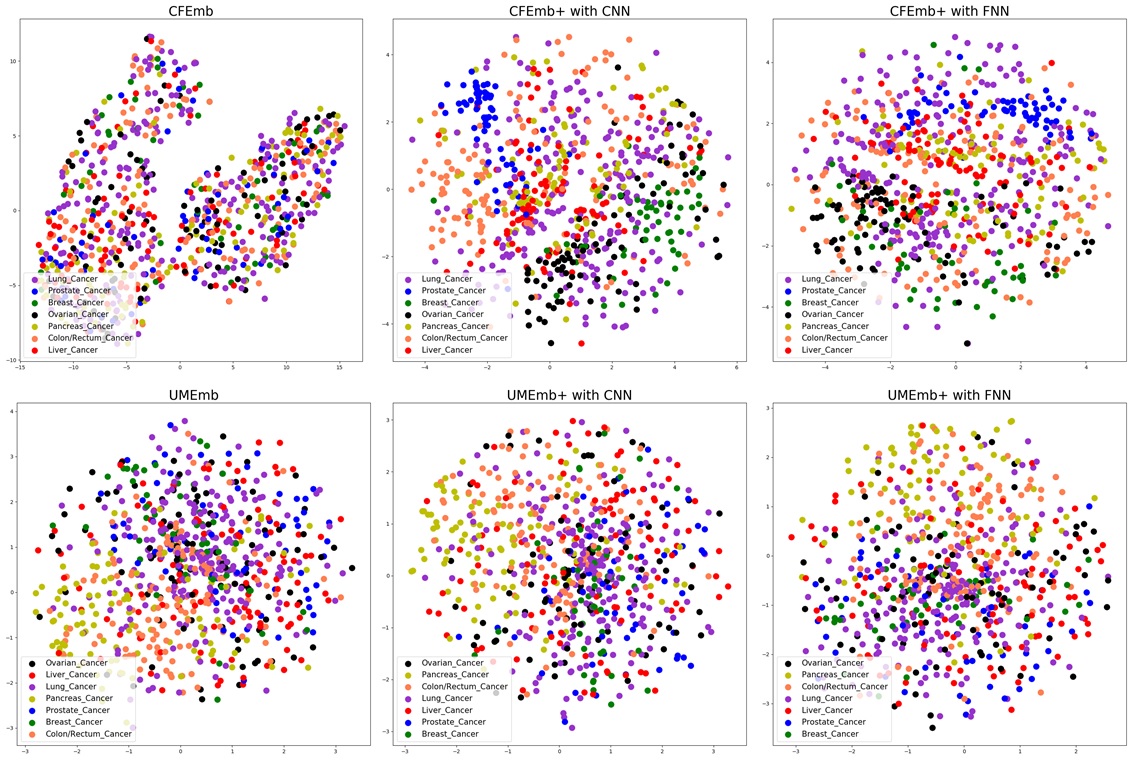}
\centering Figure 4. t-SNE clustering for different embeddings
    caption{t-SNE clustering for different embeddings}
    \label{figure4}

\end{backmatter}
\end{document}